\documentclass{article}

\PassOptionsToPackage{numbers}{natbib}

\usepackage[preprint]{neurips_2025}

\usepackage[utf8]{inputenc} 
\usepackage[T1]{fontenc}    
\usepackage{hyperref}       
\usepackage{url}            
\usepackage{booktabs}       
\usepackage{amsfonts}       
\usepackage{nicefrac}       
\usepackage{microtype}      
\usepackage[table]{xcolor}  
\usepackage{graphicx}
\usepackage{twemojis}
\usepackage{wrapfig}
\usepackage{subcaption}
\usepackage{multirow}
\usepackage{float}

\newcommand{\espresso}{\texttt{Espresso}}
\newcommand{\espressoe}{\textbf{\texttt{Espresso}} \raisebox{-0.1\ht\strutbox}{\includegraphics[height=1em]{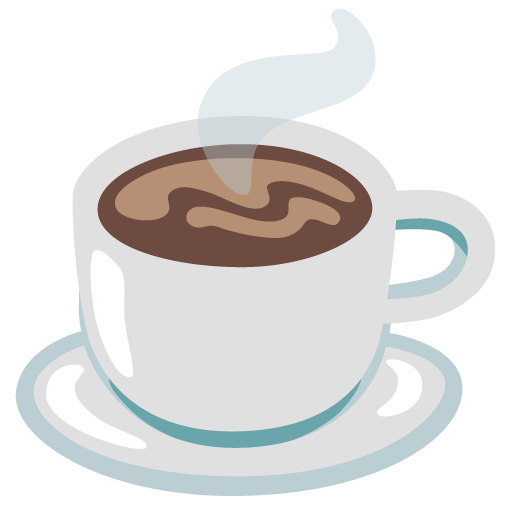}}}
\newcommand{\espressoalign}{\espresso{}\textsubscript{\texttt{align}}}
\newcommand{\espressoealign}{\espressoe{}\textsubscript{\texttt{align}}}
\newcommand{\fireemoji}{\raisebox{-0.1\ht\strutbox}{\includegraphics[height=1em]{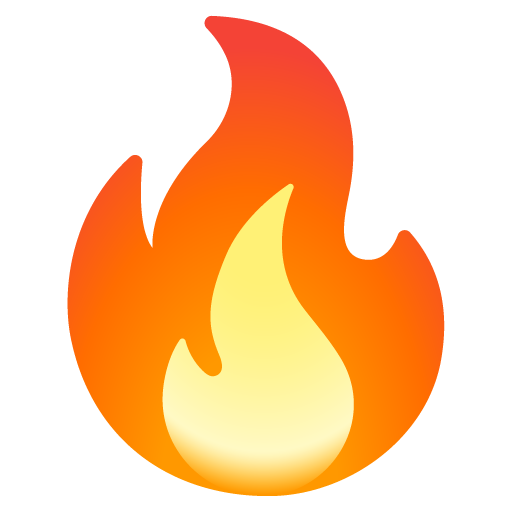}}}
\newcommand{\iceemoji}{\twemoji{2744}}
\newcommand{\withdagger}[1]{#1$^{\dagger}$}
\newcommand{\nodagger}[1]{#1\hphantom{$^{\dagger}$}}

\title{\espressoe{}: High Compression For Rich Extraction From Videos for Your Vision-Language Model}

\author{%
    Keunwoo Peter Yu\textsuperscript{1}\thanks{Corresponding author: {\tt\footnotesize kpyu@umich.edu}} \qquad
    Achal Dave\textsuperscript{2} \qquad
    Rareș Ambruș\textsuperscript{3} \qquad
    Jean Mercat\textsuperscript{3}
}

\begin{document}

\maketitle

\footnotetext[1]{University of Michigan. Work completed during an internship at Toyota Research Institute}
\footnotetext[2]{Work completed while at Toyota Research Institute}
\footnotetext[3]{Toyota Research Institute}

\begin{abstract}
Recent advances in vision-language models (VLMs) have shown great promise in connecting images and text, but extending these models to long videos remains challenging due to the rapid growth in token counts. Models that compress videos by local aggregation in time or space have become popular for handling long-form inputs; however, these pooling-based projectors sacrifice the benefits of fixed-length representations that are crucial for streaming and efficient video understanding. We introduce \espresso{}, a new architecture that separately compresses spatial and temporal features into fixed-length sequences. \espresso{} enables efficient video encoding while maintaining strong long-form reasoning capabilities. Experiments show that fixed-length compression combined with segment-wise processing offers a scalable and competitive alternative to pooling-based approaches. Our results demonstrate that fixed-length projectors, when properly designed and trained, remain a viable foundation for video-language modeling.
\end{abstract}

\section{Introduction}
\label{sec:intro}
The success of large language models (LLMs) has spurred the rise of vision-language models (VLMs) that combine visual and textual understanding by linking vision transformers (ViTs)~\cite{dosovitskiy2021an} to LLMs via a projector~\cite{alayrac2022flamingo,li2022blip,dai2023instructblip,li2023otter,awadalla2023openflamingo}. While VLMs for static images have advanced rapidly, extending these models to video presents new challenges, particularly for long-form video understanding.

The most straightforward way to adapt image VLMs to video is to treat a video as a sequence of frames~\cite{wang2024tarsier}, but this approach requires prohibitively high computational cost in practice. A ViT encodes each frame into hundreds of tokens, and processing even a handful of frames results in thousands of tokens, straining self-attention due to its quadratic complexity and overwhelming GPU memory. To address this, several VLMs use projectors that reduce the number of video tokens. Notably, Perceiver-style projectors~\cite{jaegle2021perceiver} map variable-length sequences of video tokens into fixed-length representations, enabling scalability and efficient integration with LLMs~\cite{alayrac2022flamingo,zhang-etal-2023-video,yu-etal-2024-eliciting}.

Despite their efficiency, fixed-length projectors have recently fallen out of favor. Newer models rely on pooling-based projectors that reduce tokens by a linear factor~\citep{cheng2024videollama,li2025llama-vid,laurenccon2024building}, improving general video understanding but forfeiting one of the key strengths of fixed-length methods: the ability to compress the representations of arbitrarily long videos into a constant-sized sequence. This property is especially critical in streaming and embodied settings, where models must process long sequences incrementally without overwhelming the context window of the LLM. In these scenarios, fixed-length projectors offer additional practical benefits: they keep compute and memory predictable, streamline implementation by avoiding variable-length masking, and simplify integration into low-latency inference pipelines.

In this work, we revisit the fixed-length projector paradigm and ask: \textit{Can we retain the efficiency afforded by fixed-length representations while improving their ability to understand long-form video?} Our answer is \espressoe{}---a new fixed-length projector that explicitly disentangles and compresses spatial and temporal features into separate token sequences. Inspired by architectural priors in video representation learning~\citep{simonyan2014two,feichtenhofer2019slowfast,maaz-etal-2024-video}, \espresso{} balances efficiency and expressiveness, making it well-suited for long-form video understanding.

We evaluate \espresso{} on EgoSchema~\citep{mangalam2023egoschema} and a newly proposed long-form variant, NH-EgoSchema, and find that it significantly outperforms both pooling-based and Perceiver-style projectors. We further demonstrate scalability by training on a subset of Panda-70M~\citep{chen2024panda} and the VideoChatGPT instruction tuning data~\citep{maaz-etal-2024-video}, and show strong results across a variety of video QA benchmarks. Finally, ablations reveal that separate spatial and temporal compression each contribute meaningfully to performance.

Our contributions are as follows: 1) we propose \espresso{}, a novel fixed-length projector that improves long-form video understanding by separately compressing spatial and temporal features, 2) we show that \espresso{} outperforms both pooling-based and conventional fixed-length projectors in long-form video understanding benchmarks while retaining computational efficiency, and 3) we establish that combining fixed-length projectors with segment-wise processing offers a scalable, competitive alternative to pooling-based architectures.

\section{Related Work}
\label{sec:related}

\subsection{Deep Neural Networks for Video Understanding}

\begin{wrapfigure}{r}{0.5\textwidth}
    \vspace{-13pt}
    \centering
    \includegraphics[width=\linewidth, trim= 4cm 4cm 4cm 4cm, clip]{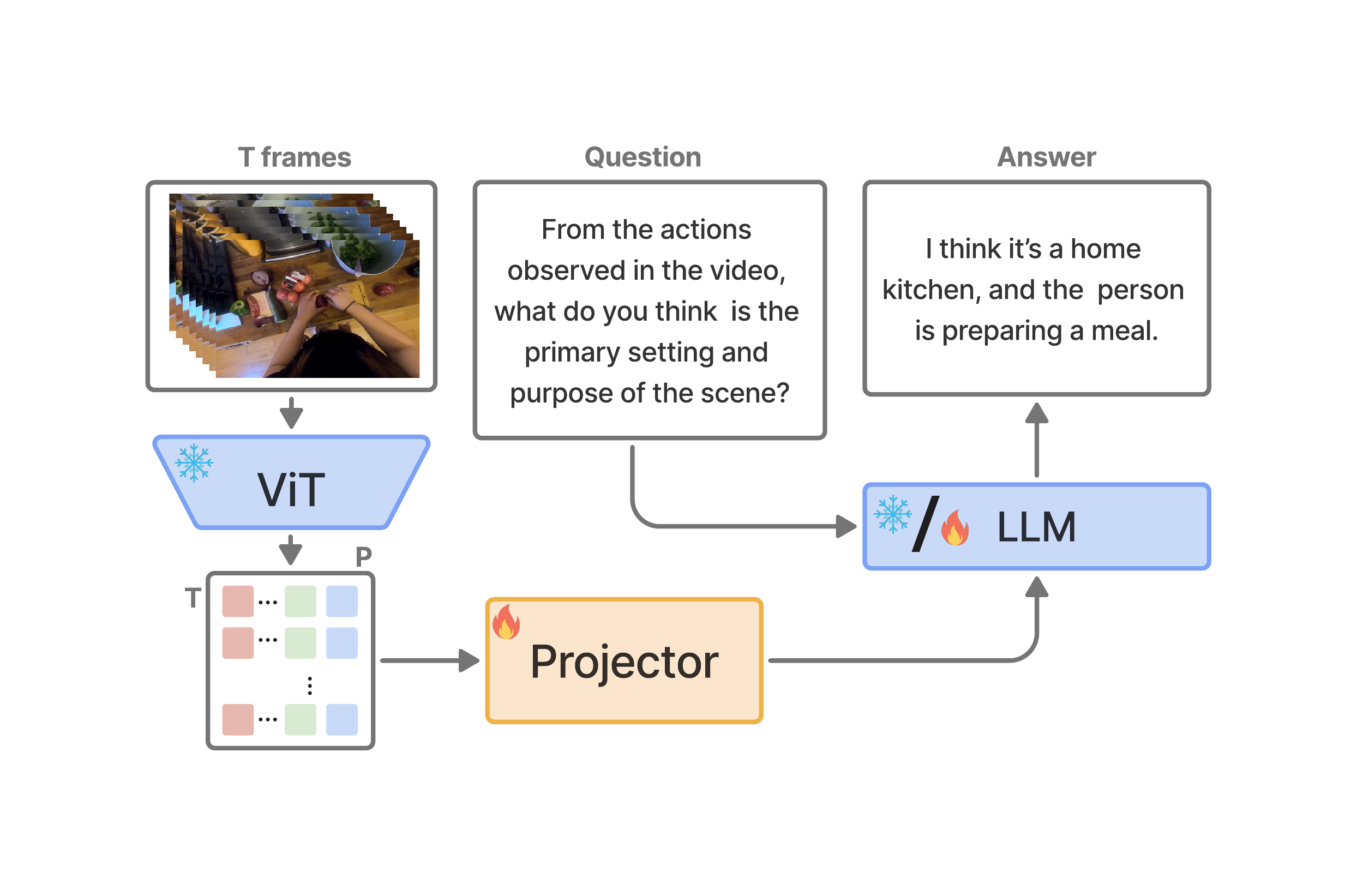}
    \caption{Overview of the canonical video language model framework. Each of the $T$ frames is first independently encoded by a frozen \iceemoji{} ViT. The frames are fed to a projector that may compress, pool or simply embed the input into the LLM input space. A pre-trained LLM, either frozen \iceemoji{} or fine-tuned \fireemoji{}, uses the projected input and a question about the video and outputs an answer.}
    \label{fig:generic_architecture}
    \vspace{-10pt}
\end{wrapfigure}

Deep neural networks for video have historically focused on action recognition. This line of research gained momentum following the success of convolutional neural networks (CNNs) in image processing. Early approaches applied CNNs directly to stacks of video frames~\citep{jhuang2009biologically,chen2010deep,le2011learning,taylor2010convolutional,ji20133d,Karpathy_2014_CVPR}, in hopes of learning spatio-temporal features implicitly. However, learning such features proved challenging~\citep{Karpathy_2014_CVPR,peng2016bag,oneata2014lear}, leading to the development of two-stream architectures that model spatial and temporal features separately~\citep{simonyan2014two,feichtenhofer2016convolutional,sevilla-lara2019on,wang2016temporal,Carreira_2017_CVPR,Sun_2018_CVPR,feichtenhofer2019slowfast}.

Similar trends have emerged with vision transformers (ViTs)~\citep{dosovitskiy2021an}, where researchers have extended transformer-based models from images to videos. These approaches fall into two main categories: joint space-time attention models~\citep{tong2022videomae,wang2023videomae,bardes2024vjepa}, which mix spatial and temporal tokens, and factorized attention models that treat space and time separately~\citep{Liu_2022_CVPR,arnab2021vivit,gberta_2021_ICML}.

\espresso{} draws inspiration from this latter group of models that disentangle spatial and temporal features, incorporating similar inductive biases into the design of fixed-length projectors for video-language modeling.




\subsection{Vision-Language Models for Video}

VLMs for video typically extend image-based VLMs by treating videos as sequences of frames (Figure~\ref{fig:generic_architecture}). Each frame is encoded by a ViT, and the resulting tokens are passed through a projector that either embeds or compresses them into the LLM input space.

\begin{figure}
    \centering
    \begin{subfigure}[t]{0.48\linewidth}
        \includegraphics[width=\linewidth]{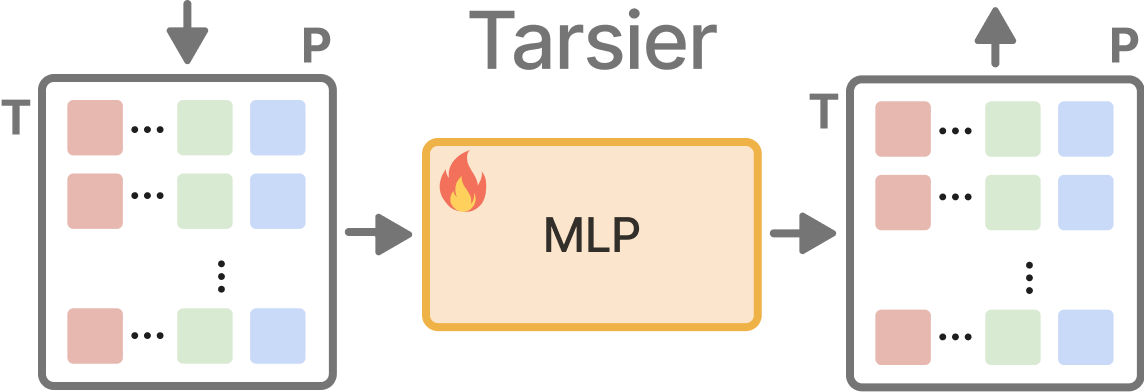}
        \caption{The projector of Tarsier~\cite{wang2024tarsier} simply projects each ViT feature with an MLP.}
        \label{fig:tarsier_architecture}
    \end{subfigure}
    \hfill
    \begin{subfigure}[t]{0.48\linewidth}
        \includegraphics[width=\linewidth, trim= 4cm 4cm 4cm 4cm, clip]{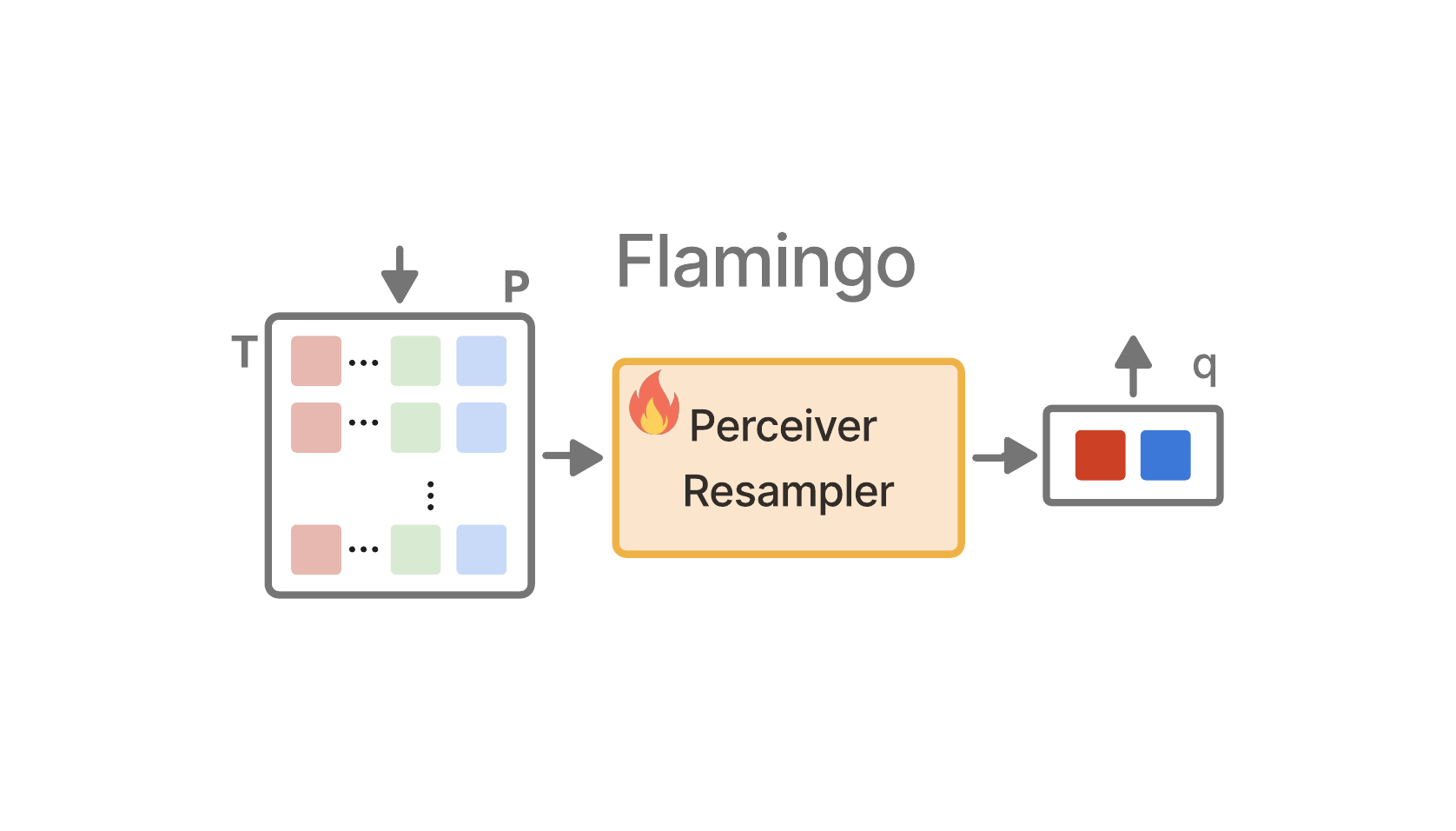}
        \caption{The projector of Flamingo~\citep{alayrac2022flamingo} uses the Perceiver Resampler to compress video tokens into fixed-length sequences.}
        \label{fig:flamingo}
    \end{subfigure}
    \begin{subfigure}[t]{0.48\linewidth}
        \includegraphics[width=\linewidth]{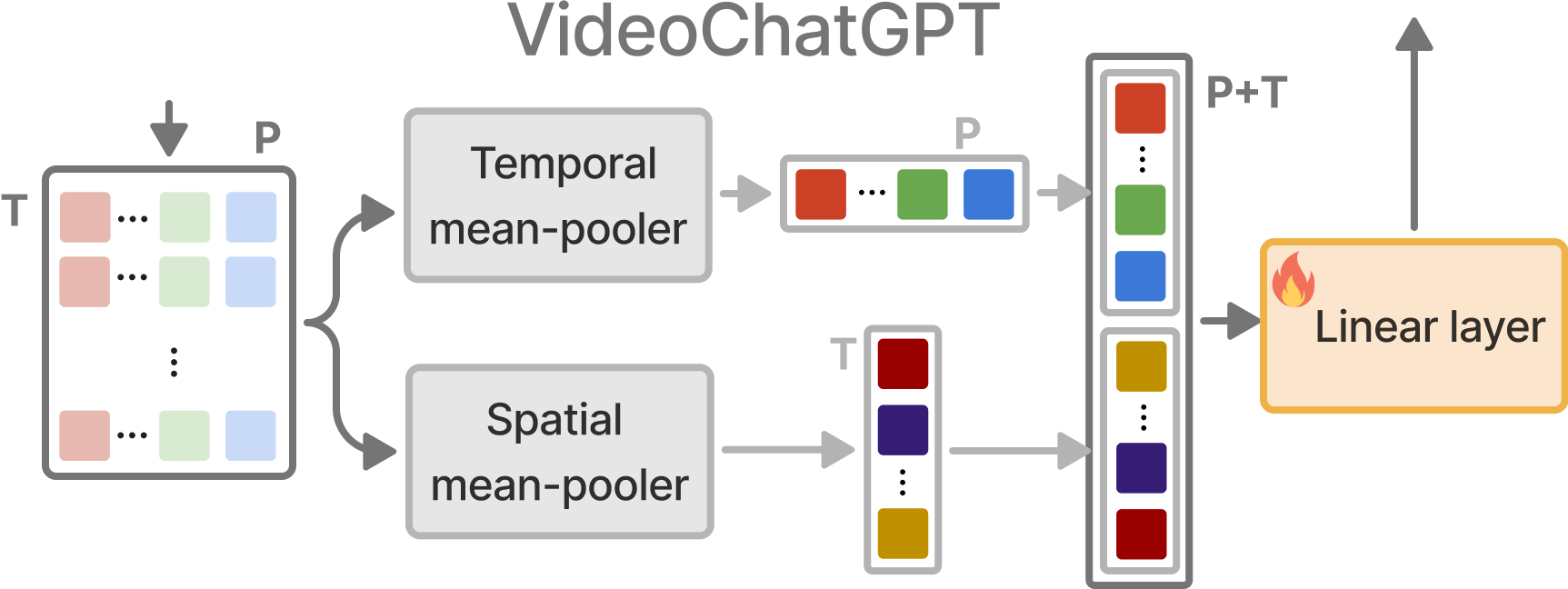}
        \caption{The projector of Video-ChatGPT~\citep{li2023videochat} uses mean pooling without any parameters and a trained multi-layer perceptron.}
        \label{fig:videochatgpt}
    \end{subfigure}
    \hfill
    \begin{subfigure}[t]{0.48\linewidth}
        \includegraphics[width=\linewidth]{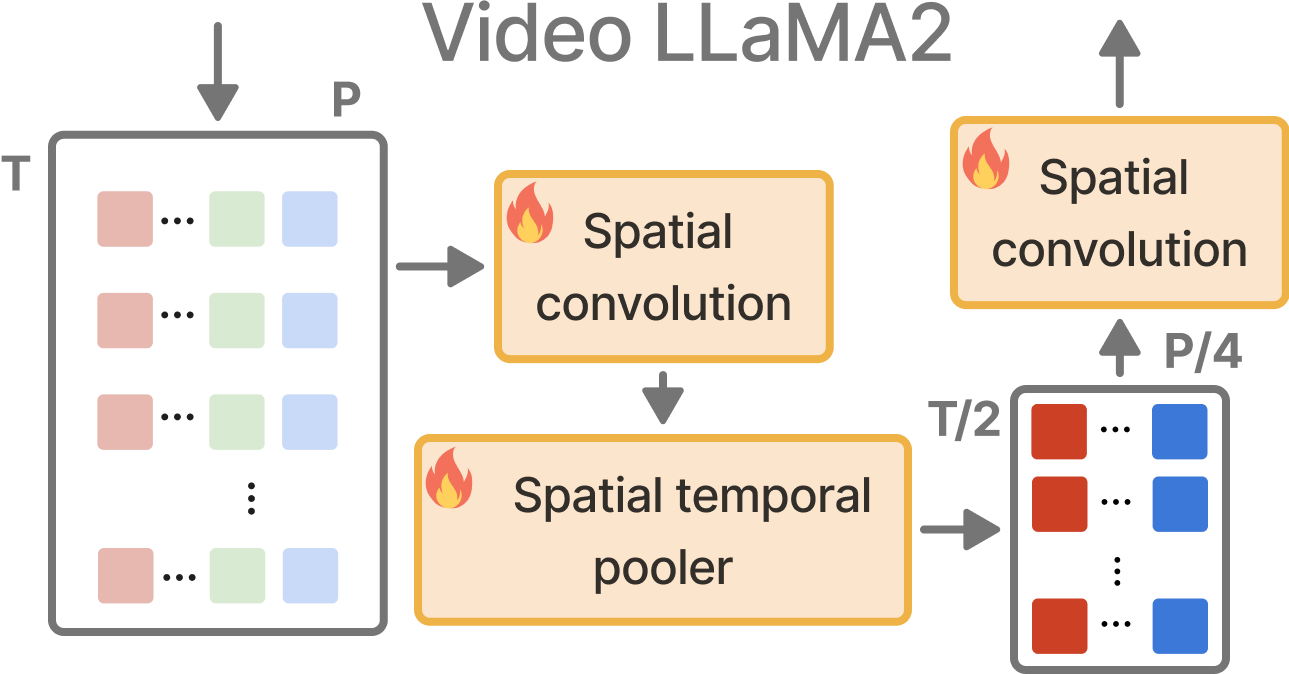}
        \caption{The projector of Video LLaMA2~\citep{cheng2024videollama} uses convolutions and pooling on spatial and temporal features.}
        \label{fig:videollama2}
    \end{subfigure}
    \caption{Overview of various projector architectures for VLMs.}
\end{figure}

\paragraph{MLP-based Projectors}  
The simplest projectors, such as those used in Tarsier~\citep{wang2024tarsier}, apply MLPs independently to each ViT token (Figure~\ref{fig:tarsier_architecture}). While straightforward, this approach results in a large number of tokens and becomes intractable for long-form video understanding.

\paragraph{Fixed-Length Projectors}  
To address scalability, several earlier works adopted Perceiver-style projectors~\citep{alayrac2022flamingo,zhang-etal-2023-video,yu-etal-2024-eliciting} that compress tokens from multiple frames into a fixed-length representation (Figure~\ref{fig:flamingo}). These models are attractive for their efficiency and ability to scale with the number of input frames. However, they have been shown to struggle with fine-grained temporal understanding, particularly in longer videos.

\espresso{} builds on this design pattern by explicitly separating and compressing spatial and temporal features into distinct fixed-length sequences. This separation allows it to preserve temporal structure more effectively while maintaining the implementation and scalability advantages of fixed-size representations---an especially important property in streaming or real-time settings.


\paragraph{Pooling-Based Projectors}  
Recent VLMs have shifted to pooling-based projectors, which reduce the number of video tokens by a linear factor while preserving temporal structure. For instance, Video-ChatGPT~\citep{maaz-etal-2024-video} applies mean pooling over frames to extract spatial and temporal features (Figure~\ref{fig:videochatgpt}), while VideoLLaMA2~\citep{cheng2024videollama} uses CNN-based downsampling (Figure~\ref{fig:videollama2}). Other approaches, such as LLaVA-Video~\citep{zhang2024video} and LLaMA-VID~\citep{li2025llama-vid}, apply various pooling or token selection strategies.

While these methods achieve strong results on general benchmarks, they sacrifice the ability to generate fixed-size video representations, which are particularly beneficial for real-time or streaming applications. In contrast, \espresso{} demonstrates that fixed-length projectors, when designed with proper architectural priors, can achieve strong long-form video understanding while retaining scalability and efficiency.

\section{Model Architecture}
\label{sec:model-arch}

In this section, we introduce the \espresso{} projector---a new fixed-length video projector that disentangles spatial and temporal features and compresses them separately into fixed-length sequences. This design enables \espresso{} to encode long-form videos efficiently without growing computational cost. Furthermore, it offers additional practical benefits such as predictable compute and memory usage, streamlined implementation by avoiding variable-length masking, and simplified integration into low-latency inference pipelines.

Most VLMs use projectors whose output scales linearly with the number of sampled frames $T$ and the number of ViT tokens per frame $P$. A representative example is the MLP-based projector in Tarsier~\citep{wang2024tarsier}, which concatenates the uncompressed ViT tokens for each frame (Figure~\ref{fig:tarsier_architecture}). This results in a token count of $O(TP)$, making it infeasible to process long videos, especially when $P$, the number of patches per frame, is in the hundreds.

\begin{figure}
    \centering
    \includegraphics[width=\linewidth, trim=4cm 4cm 4cm 4cm, clip]{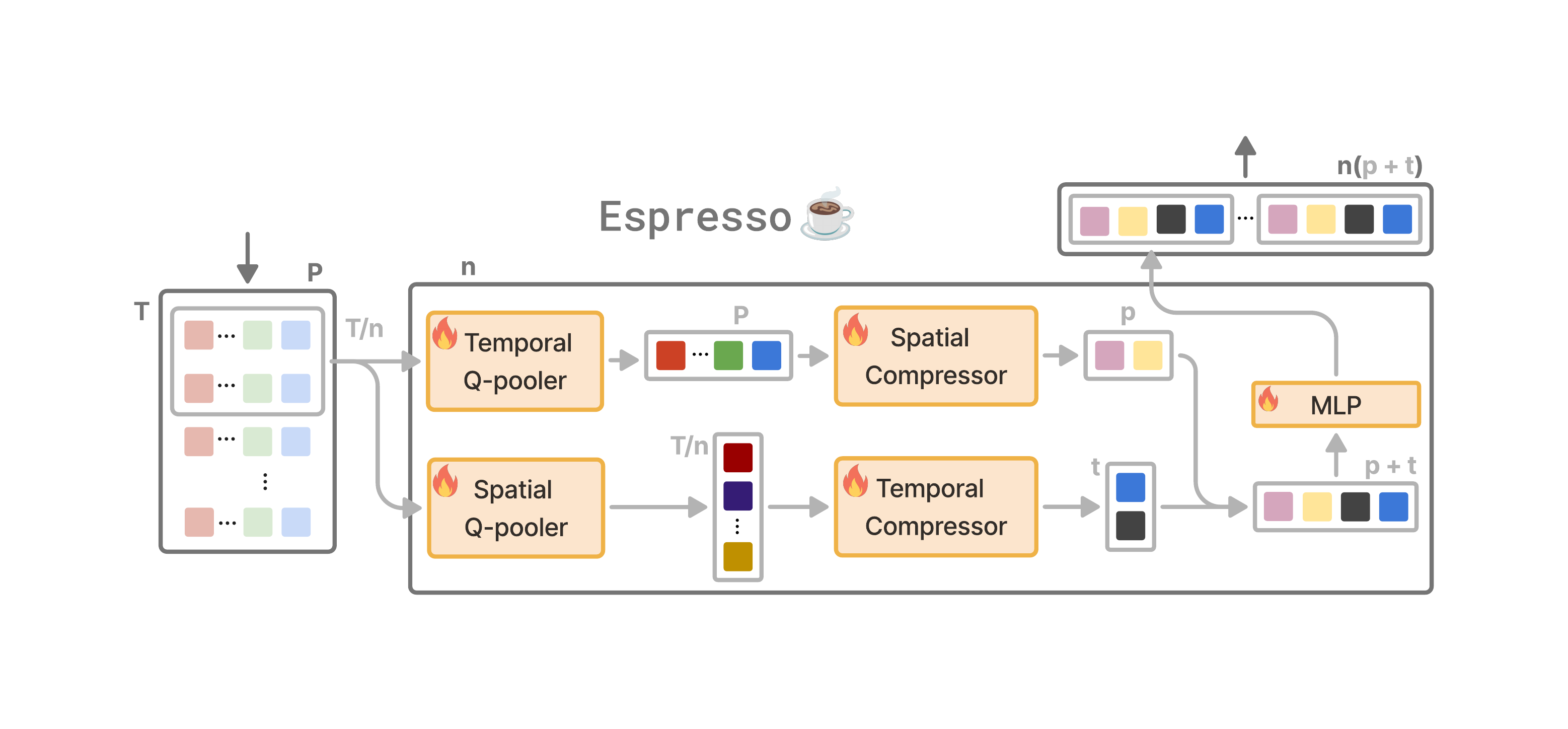}
    \caption{Overview of the \espressoe{} projector. A frozen ViT encodes each frame independently. The temporal pooler aggregates features across frames for each patch (producing spatial features), and the spatial pooler aggregates features across patches for each frame (producing temporal features). These are further compressed into fixed-length token sequences of size $p$ and $t$ respectively, then mapped into the LLM input space via an MLP. The process may optionally be repeated across $n$ segments, resulting in a token sequence of length $n(p + t)$. Note that while $P$ and $T$ are variable, $n$, $p$, and $t$ are fixed.}
    \label{fig:architecture}
    \vspace{-5pt}
\end{figure}

In contrast, \espresso{} produces a fixed-length token representation whose size does not grow with $T$ or $P$. As shown in Figure~\ref{fig:architecture}, a video $V \in \mathbb{R}^{T \times H \times W \times C}$ is independently encoded frame-by-frame by a frozen ViT to produce embeddings $X_f \in \mathbb{R}^{T \times P \times D_v}$, where $D_v$ is the ViT embedding dimension.

To extract spatial features, we rearrange $X_f$ into $X_f^t \in \mathbb{R}^{P \times T \times D_v}$ by grouping tokens from the same spatial location across time. We then apply a \textbf{temporal pooler} to this tensor, implemented as a Q-Former~\citep{li2023blip} with a single learnable query, yielding aggregated spatial features $X_s \in \mathbb{R}^{P \times D_v}$. In parallel, we extract temporal features by applying a \textbf{spatial pooler}, also a Q-Former with a single learnable query, directly to $X_f$ across the $P$ spatial locations for each of the $T$ frames. This produces $X_t \in \mathbb{R}^{T \times D_v}$.

$X_s$ and $X_t$ are independently compressed into fixed-length representations. The \textbf{spatial compressor}, implemented as a Q-Former with $p$ learnable spatial query tokens, maps $X_s$ to $\hat{X}_s \in \mathbb{R}^{p \times D_v}$. Similarly, the \textbf{temporal compressor}, also a Q-Former, maps $X_t$ to $\hat{X}_t \in \mathbb{R}^{t \times D_v}$ using $t$ learnable temporal queries. The resulting compressed features $\hat{X}_s$ and $\hat{X}_t$ are concatenated and projected into the LLM embedding space using a lightweight MLP.


We hypothesize that separating spatial and temporal compression results in representations that are more digestible for the LLM, as each compression pathway specializes in a distinct aspect of the video: the spatial path captures static scene layout, while the temporal path emphasizes motion and change. This division of labor reduces the burden on the LLM to disentangle these factors internally, enabling it to focus on higher-level reasoning. This inductive bias---explicitly modeling spatial and temporal structure via separate feature extraction and compression paths---makes the fixed-length representation more expressive than global pooling strategies, such as the Perceiver Reformer used in Flamingo~\citep{alayrac2022flamingo}.

To handle long-form videos, the input may optionally be divided into $n$ temporal segments, each of which is processed independently following the same scheme. This yields $n$ separate compressed representations of size $(p + t)$ tokens each, which are concatenated and passed to the LLM. The total input length to the LLM is thus fixed at $n(p + t)$, making \espresso{} particularly well-suited for applications that involve streaming, sliding windows, or strict context length budgets.

In contrast to pooling-based projectors, which reduce token count but grow linearly with $T$, \espresso{} preserves a fixed-size representation and achieves strong long-form video understanding---a property we validate through extensive experiments.

\section{Evaluation}
\label{sec:eval}

We conduct extensive evaluations to validate the effectiveness of \espresso{}. First, we perform controlled experiments to assess its long-form video understanding capabilities relative to other projector designs, while matching or exceeding their computational efficiency. We then scale up training and evaluate \espresso{} across a range of video understanding benchmarks.

\subsection{Long-Form Video Understanding with Different Projectors}
\label{subsec:long-form-vid-understanding-projectors}

We begin with tightly controlled experiments to evaluate whether \espresso{} achieves stronger long-form video understanding than other popular projectors. We compare the following methods:

\begin{itemize}
    \item \textbf{MLP}: the non-compressive, MLP-based projector from Tarsier~\citep{wang2024tarsier}
    \item \textbf{STC}: the Spatial-Temporal Convolution projector used in VideoLLaMA2~\citep{cheng2024videollama}, which compresses frame features using RegStage~\citep{Radosavovic_2020_CVPR} and convolutional downsampling
    \item \textbf{PR}: a fixed-length projector with global pooling, implemented using a single Q-Former in the spirit of the Perceiver Reformer from Flamingo~\citep{alayrac2022flamingo}
\end{itemize}

All models are initialized from the same pretrained 7B LLaVA checkpoint from Prismatic~\citep{karamcheti2024prismatic}. For consistency, we freeze both the ViT and LLM and train only the projector and the final MLP using the Video-ChatGPT instruction tuning dataset~\citep{maaz-etal-2024-video} for 3 epochs with a batch size of 32 and a learning rate of 2e-5. We disable \espresso{}’s segmenting mechanism (i.e., treating the input as a single continuous sequence) to match the input structure of the other projectors and fix the number of spatial and temporal tokens to $p = t = 4$. The effects of the different numbers of spatial and temporal tokens are investigated in Section~\ref{sec:ablation}. To ensure fair comparison, we also set the number of query tokens for PR to 8, matching the total number of learnable tokens used by \espresso{}.

We use EgoSchema~\citep{mangalam2023egoschema} as our primary evaluation benchmark due to its reproducibility (multiple-choice QA format rather than free-form generation) and its explicit emphasis on long-form video understanding. To further stress long-range reasoning, we introduce a more challenging variant: \textbf{NH-EgoSchema} (needle-in-a-haystack EgoSchema). In NH-EgoSchema, each evaluation video is constructed by randomly sampling and concatenating four EgoSchema videos---one target and three distractors---then randomly shuffling their order. The resulting composite video is twelve minutes long, and the prompt specifies which scene the question is about.

\begin{table}
\centering
\caption{Long-form video understanding evaluation. \textbf{Default} refers to EgoSchema’s standard setup; \textbf{NH} refers to our proposed needle-in-a-haystack variant. MLP fails to train on 64 and 128 frames due to memory limits (even on an A100 80GB GPU). Best results in each setting are bolded.}
\label{tab:projector}
\begin{tabular}{@{}ccccccccc@{}}
\toprule
Model & \multicolumn{2}{c}{\espressoe} & \multicolumn{2}{c}{MLP} & \multicolumn{2}{c}{STC} & \multicolumn{2}{c}{PR} \\ \midrule
Frames & Default        & NH             & Default         & NH    & Default & NH    & Default & NH    \\ \midrule
8      & 37.87          & \textbf{32.06} & \textbf{39.18}  & 30.89 & 37.13   & 30.65 & 38.22   & 30.39 \\
16     & \textbf{42.18} & \textbf{35.86} & 31.62           & 28.17 & 34.59   & 29.60 & 32.58   & 28.11 \\
32     & \textbf{43.35} & \textbf{33.85} & 25.92           & 23.61 & 31.58   & 26.28 & 34.35   & 29.46 \\
64     & \textbf{40.85} & \textbf{32.78} & \multicolumn{2}{c}{OOM} & 27.25   & 24.47 & 33.29   & 27.47 \\
128    & \textbf{40.17} & \textbf{32.98} & \multicolumn{2}{c}{OOM} & 23.10   & 23.18 & 36.08   & 28.64 \\ \bottomrule
\end{tabular}
\end{table}

As shown in Table~\ref{tab:projector}, \espresso{} consistently outperforms all baselines on both EgoSchema and NH-EgoSchema across all frame counts beyond 8. Notably, only the fixed-length projectors---\espresso{} and PR---show improved performance as the number of sampled frames increases. However, \espresso{} outperforms PR in all settings except for 8 frames, highlighting the effectiveness of its spatial-temporal compression strategy over global pooling.

These results support our central hypothesis: by explicitly disentangling and compressing spatial and temporal features into fixed-length sequences, \espresso{} produces structured representations that are not only efficient but also more digestible for LLMs—leading to improved long-form video understanding.

\subsection{Computational Efficiency}

At first glance, \espresso{} may not appear computationally efficient due to its relatively large parameter count---it uses four Q-Formers, compared to one in PR and none in MLP. However, the primary computational bottleneck in VLMs is not in the projector, but in the LLM backbone, which dominates both the parameter count and runtime. As such, inference efficiency is more strongly influenced by the number of tokens fed into the LLM. In this regard, fixed-length projectors like \espresso{} excel, as they keep the LLM input length constant regardless of video length.

To validate this claim, we measure inference time and peak memory usage for each VLM evaluated in Section~\ref{subsec:long-form-vid-understanding-projectors}. We use the same prompt and video across models, varying only the number of sampled frames. For each setting, we perform two warm-up runs followed by 10 evaluation runs. Each run generates a single output token, and we report the average CUDA runtime and peak memory usage across the 10 runs. All models are loaded in \texttt{bfloat16} precision and executed on a single Quadro RTX 6000 GPU with 24GB of memory.

\begin{table}
\centering
\caption{Computational efficiency evaluation. \textbf{Time} indicates average CUDA time per generation step; \textbf{Mem} is peak memory usage; \textbf{Param} is the number of parameters in each projector module (not counting the ViT or LLM).}
\label{tab:time-mem}
\begin{tabular}{@{}ccccccccc@{}}
\toprule
Model & \multicolumn{2}{c}{\espressoe} & \multicolumn{2}{c}{MLP}   & \multicolumn{2}{c}{STC}   & \multicolumn{2}{c}{PR}     \\ \midrule
Param & \multicolumn{2}{c}{726.4M}     & \multicolumn{2}{c}{21.0M} & \multicolumn{2}{c}{50.5M} & \multicolumn{2}{c}{197.3M} \\ \midrule
Frames & Time & Mem   & Time       & Mem        & Time & Mem   & Time & Mem   \\ \midrule
8      & .32s & 15.2G & .13s       & 18.0G      & .32s & 14.4G & .23s & 14.2G \\
16     & .39s & 15.2G & \multicolumn{2}{c}{OOM} & .51s & 14.9G & .29s & 14.2G \\
32     & .52s & 15.6G & \multicolumn{2}{c}{OOM} & .89s & 16.0G & .41s & 14.6G \\
64     & .76s & 16.4G & \multicolumn{2}{c}{OOM} & 1.7s & 18.4G & .66s & 15.4G \\
128   & 1.3s          & 17.9G          & \multicolumn{2}{c}{OOM}   & \multicolumn{2}{c}{OOM}   & 1.2s        & 16.9G        \\ \bottomrule
\end{tabular}
\end{table}

Table~\ref{tab:time-mem} presents the results. As expected, the fixed-length projectors---\espresso{} and PR---scale best as the number of sampled frames increases, despite having significantly larger parameter counts. By contrast, the smallest projector (MLP) performs worst in terms of memory efficiency and runs out of memory at 16 frames on a 24GB GPU. However, it achieves the fastest inference time at 8 frames. This is because, at lower frame counts, the efficiency gains from token compression are insufficient to offset the additional compute overhead of the compressive projectors. As a result, a simple non-compressive projector like MLP can perform well. STC maintains efficiency at low frame counts but quickly degrades with longer inputs.

PR achieves the highest computational efficiency among the fixed-length projectors, as it compresses the entire video using a single Q-Former. However, as shown in Section~\ref{subsec:long-form-vid-understanding-projectors}, PR consistently underperforms \espresso{} on long-form benchmarks. This highlights a key trade-off: while both projectors benefit from fixed-size representations, \espresso{}'s explicit spatial-temporal disentanglement leads to richer representations and stronger long-range video reasoning.

\subsection{Video Understanding Benchmarks}

Having validated \espresso{}'s long-form video understanding capabilities and computational efficiency, we next evaluate its performance across a broader range of video-language benchmarks. In particular, we assess whether fixed-length projectors can remain competitive on both short-form and long-form tasks.

We organize our evaluations into two categories:
\begin{itemize}
    \item \textbf{Short-Form Video Understanding}: Benchmarks where reasoning over a few frames typically suffices, such as MSVD-QA~\citep{xu2017video}, MSRVTT-QA~\citep{xu2017video}, TGIF-QA~\citep{jang2017tgif}, ActivityNet-QA~\citep{yu2019activitynet}, Video-ChatGPT~\citep{maaz-etal-2024-video}, and MovieChat-1K~\citep{Song_2024_CVPR}
    \item \textbf{Long-Form Video Understanding}: Benchmarks that require sustained temporal reasoning over long video sequences, including EgoSchema~\citep{mangalam2023egoschema}, MLVU~\citep{zhou2024mlvu}, and Video-MME~\citep{fu2024video}
\end{itemize}

We consider two variants of our model:
\begin{itemize}
    \item \espresso{}: trained only on the Video-ChatGPT instruction tuning dataset with about 100k samples, using 128 sampled frames divided into 8 segments (16 frames per segment)
    \item \espressoalign{}: first trained on a 14M-sample subset of Panda-70M~\citep{chen2024panda} without segments (16 global frames), then fine-tuned on the Video-ChatGPT instruction tuning dataset with segments. This two-stage training strategy aims to bootstrap projector alignment using abundant short-form video-caption pairs before specializing on instruction-following with long videos
\end{itemize}

Both variants are initialized from the same 7B LLaVA checkpoint from Prismatic~\citep{karamcheti2024prismatic}. We train only the \espresso{} projector while freezing the ViT and LLM, as unfreezing the LLM backbone degraded performance, likely due to the limited diversity and small size of the instruction tuning data.

As baselines, we use the following models:
\begin{itemize}
    \item \textbf{VideoLLaMA2~\citep{cheng2024videollama}}: A VLM that uses the pooling-based STC projector. It is trained on over 15 million video-, image-, and text-based examples from a mix of public and private sources.
    \item \textbf{Tarsier~\citep{wang2024tarsier}}: A VLM using a simple MLP projector, trained on 6 million video-text pairs and 500k in-house instruction-tuning examples.
    \item \textbf{Video-LLaVA~\citep{lin-etal-2024-video}}: A VLM that incorporates LanguageBind~\citep{zhu2024languagebind} as its fixed-length projector. It is trained on a mixture of roughly 2 million image-text and video-text pairs, including instruction-tuning data.
    \item \textbf{MovieChat~\citep{Song_2024_CVPR}}: A VLM that combines an MLP-based projector for short-term memory with a Q-Former-based fixed-length projector for long-term memory. It is trained on MovieChat-1K, a curated dataset of densely annotated movie and TV clips.
\end{itemize}

Note that these models are trained on larger and more diverse datasets than those used for \espresso{}. While \espressoalign{} is trained on a larger dataset due to the addition of a subset of Panda-70M, its overall data diversity remains much lower.

\subsubsection{Short-Form Video Understanding}
We first evaluate \espresso{} and \espressoalign{} on benchmarks where videos are relatively short and global context can often be captured within a few frames. These include MSVD-QA~\citep{xu2017video}, MSRVTT-QA~\citep{xu2017video}, TGIF-QA~\citep{jang2017tgif}, ActivityNet-QA~\citep{yu2019activitynet}, and the Video-ChatGPT evaluation suite~\citep{maaz-etal-2024-video}. We also assess performance on MovieChat-1K~\citep{Song_2024_CVPR}, which consists of long movie clips but often requires reasoning over localized events.

\begin{table}[t]
\centering
\caption{Zero-shot video QA results. All models use 7B LLM backbones. \textbf{Acc} = accuracy, \textbf{Sco} = score. The first frame number for \espresso{} models indicates total frames, the second number indicates frames per segment. Best results in bold; $\dag$ results are official; rest are reproduced by us.}
\label{tab:short-video-understanding}
\begin{tabular}{@{}lc|cccccccc@{}}
\toprule
\multirow{2}{*}{Model}                 & \multirow{2}{*}{Frames} & \multicolumn{2}{c}{MSVD-QA}                            & \multicolumn{2}{c}{MSRVTT-QA}                          & \multicolumn{2}{c}{TGIF-QA}                        & \multicolumn{2}{c}{ActivityNet-QA}                     \\ \cmidrule(l){3-10} 
                                       &                         & Acc                        & Sco                       & Acc                        & Sco                       & Acc                      & Sco                     & Acc                        & Sco                       \\ \midrule
VideoLLaMA2~\cite{cheng2024videollama} & 16                      & \withdagger{70.6}          & \withdagger{3.8}          & \nodagger{55.0}            & \nodagger{3.4}            & \textbf{\nodagger{80.8}} & \textbf{\nodagger{4.4}} & \withdagger{53.0}          & \withdagger{3.4}          \\
Video-LLaVA~\cite{lin-etal-2024-video} & 8                       & \withdagger{70.7}          & \withdagger{3.9}          & \withdagger{59.2}          & \textbf{\withdagger{3.5}} & \withdagger{70.0}        & \withdagger{4.0}        & \withdagger{45.3}          & \withdagger{3.3}          \\
MovieChat~\cite{Song_2024_CVPR}        & 2048                    & \withdagger{75.2}          & \withdagger{3.8}          & \withdagger{52.7}          & \withdagger{2.6}          & \nodagger{-}             & \nodagger{-}            & \withdagger{45.7}          & \withdagger{3.4}          \\
Tarsier~\cite{wang2024tarsier}         & 16                      & \textbf{\withdagger{77.0}} & \textbf{\withdagger{4.1}} & \textbf{\withdagger{62.0}} & \textbf{\withdagger{3.5}} & \withdagger{79.2}        & \withdagger{4.2}        & \textbf{\withdagger{59.5}} & \textbf{\withdagger{3.6}} \\ \midrule
\espressoe                             & 128/16                  & \nodagger{64.8}            & \nodagger{3.8}            & \nodagger{55.0}            & \nodagger{3.4}            & \nodagger{51.8}          & \nodagger{3.4}          & \nodagger{40.3}            & \nodagger{3.2}            \\ \midrule
\espressoealign                        & 128/16                  & \nodagger{72.0}            & \nodagger{4.0}            & \nodagger{57.3}            & \nodagger{3.5}            & \nodagger{71.3}          & \nodagger{4.0}          & \nodagger{44.3}            & \nodagger{3.4}            \\
\espressoealign                        & 64/16                   & \nodagger{71.3}            & \nodagger{4.0}            & \nodagger{58.1}            & \nodagger{3.5}            & \nodagger{71.2}          & \nodagger{4.0}          & \nodagger{44.4}            & \nodagger{3.4}            \\
\espressoealign                        & 32/16                   & \nodagger{71.1}            & \nodagger{4.0}            & \nodagger{57.5}            & \nodagger{3.5}            & \nodagger{71.4}          & \nodagger{4.0}          & \nodagger{43.4}            & \nodagger{3.3}            \\
\espressoealign                        & 16/16                   & \nodagger{71.6}            & \nodagger{4.0}            & \nodagger{57.5}            & \nodagger{3.5}            & \nodagger{71.9}          & \nodagger{4.0}          & \nodagger{42.9}            & \nodagger{3.3}            \\ \bottomrule
\end{tabular}
\end{table}

\begin{table}[t]
\centering
\caption{Video-ChatGPT evaluation results. Metrics: \textbf{Cor} (Correctness), \textbf{D} (Detail), \textbf{Ctx} (Context), \textbf{T} (Temporal), \textbf{Con} (Consistency). All models use 7B LLM backbones. First/second frame numbers indicate total frames/frames per segment. Best results in bold; $\dag$ results are official; rest are reproduced by us.}
\label{tab:video-chatgpt-eval}
\begin{tabular}{@{}lc|ccccc@{}}
\toprule
Model                                  & Frames & Cor                     & D                         & Ctx                       & T                         & Con                       \\ \midrule
VideoLLaMA2~\cite{cheng2024videollama} & 16     & \withdagger{3.3}        & \textbf{\withdagger{3.2}} & \textbf{\withdagger{3.8}} & \textbf{\withdagger{2.7}} & \textbf{\withdagger{3.2}} \\
MovieChat~\cite{Song_2024_CVPR}        & 2048   & \withdagger{2.8}        & \withdagger{2.9}          & \withdagger{3.0}          & \withdagger{2.2}          & \withdagger{2.4}          \\
Tarsier~\cite{wang2024tarsier}         & 16     & \textbf{\nodagger{3.4}} & \nodagger{3.1}            & \nodagger{3.7}            & \textbf{\nodagger{2.7}}   & \nodagger{1.2}            \\ \midrule
\espressoe                             & 128/16 & \nodagger{2.6}          & \nodagger{2.4}            & \nodagger{3.0}            & \nodagger{2.3}            & \nodagger{2.4}            \\ \midrule
\espressoealign                        & 128/16 & \nodagger{3.0}          & \nodagger{2.6}            & \nodagger{3.3}            & \nodagger{2.3}            & \nodagger{2.8}            \\
\espressoealign                        & 64/16  & \nodagger{2.9}          & \nodagger{2.6}            & \nodagger{3.3}            & \nodagger{2.4}            & \nodagger{2.8}            \\
\espressoealign                        & 32/16  & \nodagger{2.9}          & \nodagger{2.6}            & \nodagger{3.3}            & \nodagger{2.3}            & \nodagger{2.8}            \\
\espressoealign                        & 16/16  & \nodagger{2.8}          & \nodagger{2.5}            & \nodagger{3.2}            & \nodagger{2.3}            & \nodagger{2.7}            \\ \bottomrule
\end{tabular}
\end{table}

\begin{table}
\centering
\caption{MovieChat-1K results. \textbf{Acc} = accuracy, \textbf{Sco} = score. All models use 7B LLM backbones. First/second frame numbers indicate total frames/frames per segment. Best results in bold.}
\label{tab:moviechat-eval}
\begin{tabular}{@{}lc|cccc@{}}
\toprule
\multirow{2}{*}{Model}                 & \multirow{2}{*}{Frames} & \multicolumn{2}{c}{Global}   & \multicolumn{2}{c}{Breakpoint} \\ \cmidrule(l){3-6} 
                                       &                         & Acc           & Sco          & Acc            & Sco           \\ \midrule
VideoLLaMA2~\cite{cheng2024videollama} & 16                      & 70.2          & 3.9          & 64.1           & 3.6           \\
Tarsier~\cite{wang2024tarsier}         & 16                      & \textbf{76.7} & \textbf{4.0} & \textbf{71.2}  & \textbf{3.8}  \\ \midrule
\espressoe                             & 128/16                  & 48.8          & 3.4          & 40.2           & 2.9           \\ \midrule
\espressoealign                        & 128/16                  & 67.8          & 3.8          & 58.4           & 3.4           \\
\espressoealign                        & 64/16                   & 64.5          & 3.8          & 58.0           & 3.4           \\
\espressoealign                        & 32/16                   & 67.8          & 3.8          & 58.3           & 3.5           \\
\espressoealign                        & 16/16                   & 64.7          & 3.8          & 56.0           & 3.4           \\ \bottomrule
\end{tabular}
\end{table}

Results are shown in Tables \ref{tab:short-video-understanding}, \ref{tab:video-chatgpt-eval}, and \ref{tab:moviechat-eval}. Across these benchmarks, \espressoalign{} outperforms \espresso{} by a significant margin, demonstrating that large-scale projector alignment on Panda-70M improves performance on short-form video understanding. The improvements are consistent across multiple segment settings, suggesting that \espressoalign{} learns to extract salient features even in settings where explicit segmenting is unnecessary, e.g., short-form video understanding tasks.

Although \espressoalign{} does not surpass the best-performing baselines (e.g., Tarsier~\citep{wang2024tarsier} and VideoLLaMA2~\citep{cheng2024videollama}), its performance is competitive given its lower-diversity training data and reliance on a fixed-length compression strategy rather than pooling or naive token concatenation. Notably, segment-based processing---while critical for long-form video understanding---offers little advantage on these short-form benchmarks. This aligns with intuition: when videos are short and questions require only a small number of frames to answer, the benefits of aggressive compression and segment-wise processing diminish.

These results suggest that fixed-length projectors like \espresso{} remain viable even when long-form temporal reasoning is not explicitly required. Moreover, their performance on short-form tasks can be further improved through scalable two-stage training strategies without needing to retrain the entire model.

\subsubsection{Long-Form Video Understanding}
We now turn to benchmarks that explicitly test sustained reasoning over long videos: EgoSchema~\citep{mangalam2023egoschema}, its needle-in-a-haystack variant (NH-EgoSchema), MLVU~\citep{zhou2024mlvu}, MovieChat-1K~\citep{Song_2024_CVPR}, and Video-MME~\citep{fu2024video}.

\begin{table}
\centering
\caption{Long-form video understanding results. Metrics: \textbf{M-Avg} (MLVU multiple-choice avg), \textbf{Default} (EgoSchema standard), \textbf{NH} (needle-in-a-haystack variant), \textbf{Short/Med/Long} (VideoMME splits based on video lengths). Frame numbers indicate total frames/frames per segment. Best results in bold. All results are reproduced.}
\label{tab:long-form-eval}
\begin{tabular}{@{}lc|c|cc|cccc@{}}
\toprule
\multirow{3}{*}{Model}                 & \multirow{3}{*}{Frames} & \multirow{2}{*}{MLVU} & \multicolumn{2}{c|}{\multirow{2}{*}{EgoSchema}} & \multicolumn{4}{c}{\multirow{2}{*}{VideoMME}}                 \\
                                       &                         &                       & \multicolumn{2}{c|}{}                           & \multicolumn{4}{c}{}                                          \\
                                       &                         & M-Avg                 & Default                & NH                     & Short         & Med           & Long          & Overall       \\ \midrule
VideoLLaMA2~\cite{cheng2024videollama} & 16                      & \textbf{45.6}         & \textbf{53.1}          & 43.3                   & \textbf{61.4} & \textbf{53.4} & \textbf{47.3} & \textbf{54.1} \\
Tarsier~\cite{wang2024tarsier}         & 16                      & 32.4                  & 49.9                   & 37.5                   & 48.9          & 37.6          & 32.9          & 39.8          \\ \midrule
\espressoe                             & 128/16                  & 31.2                  & 50.4                   & \textbf{45.1}          & 34.7          & 32.9          & 32.8          & 33.4          \\ \midrule
\espressoealign                        & 128/16                  & 34.5                  & 42.1                   & 32.5                   & 43.8          & 36.7          & 34.9          & 38.4          \\
\espressoealign                        & 64/16                   & 33.2                  & 41.5                   & 33.7                   & 42.1          & 36.8          & 34.0          & 37.6          \\
\espressoealign                        & 32/16                   & 30.3                  & 43.4                   & 36.0                   & 41.7          & 35.1          & 35.0          & 37.3          \\
\espressoealign                        & 16/16                   & 29.2                  & 39.5                   & 30.5                   & 40.7          & 35.9          & 34.3          & 37.0          \\ \bottomrule
\end{tabular}
\end{table}

Results are shown in Tables~\ref{tab:long-form-eval}. Several key patterns emerge:

\begin{itemize}
    \item \textbf{Segments matter.} Unlike short-form tasks, segment-wise processing consistently improves performance on long-form benchmarks, highlighting the importance of localized temporal compression for scaling to longer videos.
    \item \textbf{\espresso{} performs robustly on challenging long-form tasks.} On NH-EgoSchema, the most difficult evaluation setting, \espresso{} outperforms all baselines, including the pooling-based model like VideoLLaMA2. It also modestly surpasses Tarsier on the standard EgoSchema benchmark, though it trails VideoLLaMA2. These two benchmarks emphasize long-form temporal reasoning, making them especially sensitive to how well models capture extended temporal dependencies. While \espressoalign{} generally performs better on other benchmarks, it underperforms \espresso{} here---likely due to its pretraining on short video-caption pairs, which may bias it toward local patterns. In contrast, \espresso{}'s direct fine-tuning on segmented long videos helps it better model long-range structure.    
    \item \textbf{\espressoalign{} generalizes better across diverse long videos.} On broader benchmarks like MLVU and Video-MME, \espressoalign{} closes the gap with, or outperforms, Tarsier despite much less diverse training data. Its two-stage training procedure appears to yield more general-purpose video representations, even if it slightly underperforms \espresso{} on more focused, pure long-form tasks like EgoSchema and NH-EgoSchema.
    \item \textbf{Pooling baselines still have advantages---but at a cost.} While pooling-based architectures like VideoLLaMA2 achieve higher absolute scores, they sacrifice fixed-size scalability. Our results suggest that with further improvements in training data and segment-aware pretraining, fixed-length projectors like \espresso{} could become even more competitive for long-form video understanding.
\end{itemize}

Overall, these results validate our central claim: fixed-length projectors remain a viable architecture for long-form video-language modeling, offering competitive efficiency and scalability. While there is still a gap to the strongest pooling-based models, especially those trained with massive curated datasets, our findings highlight the untapped potential of fixed-length compression strategies when combined with appropriate training designs.

\section{Conclusion}

In this work, we revisit fixed-length projectors for VLMs for video, motivated by their desirable properties for long-form and streaming settings. We introduce \espresso{}, a fixed-length projector that separately compresses spatial and temporal features, and demonstrate its strong performance across diverse video understanding benchmarks. Through extensive evaluations and ablations, we show that fixed-length compression combined with segment-wise processing offers a scalable and competitive alternative to pooling-based architectures. These findings underscore the promise of fixed-length projectors for efficient and effective long-form video understanding, and pave the way for further exploration in using fixed-length projectors for streaming and embodied settings.

\paragraph{Acknowledgments}
We thank Thomas Kollar for his valuable insights in shaping the direction of this research.


\bibliographystyle{plain}
\bibliography{references}


\appendix

\section{Ablation Studies}
\label{sec:ablation}

We now conduct a series of ablation studies to better understand how spatial and temporal compression, as well as the use of temporal segments, influence long-form video understanding in \espresso{}. All experiments in this section follow the same setup as in Section~\ref{subsec:long-form-vid-understanding-projectors}. Specifically, we initialize the ViT, LLM, and MLP from Prismatic’s pretrained LLaVA checkpoint~\citep{karamcheti2024prismatic}, fine-tune only the projector on instruction-tuning data from Video-ChatGPT~\citep{maaz-etal-2024-video}, and evaluate on EgoSchema~\citep{mangalam2023egoschema} using its default setup.

\subsection{Spatial Compression}
\label{subsec:spatial-compression}

We first examine how spatial compression affects downstream performance. We fix the number of sampled frames to 32, disable the temporal compressor, and vary $p$, the number of spatial query tokens, isolating the impact of spatial compression. We define the spatial compression rate as $-\log(p)$. A higher compression rate corresponds to fewer spatial tokens.

\begin{table}
\centering
\caption{EgoSchema accuracy for different spatial compression rates in the default setting. The video is sampled at 32 frames and evenly divided into segments. $p$ is the number of spatial query tokens. Best accuracy for each column is in bold. Gray cells denote no spatial compression. $r$ is the Pearson correlation coefficient between spatial compression rate ($-\log(p)$) and model accuracy.}
\label{tab:spatial-compression}
\begin{tabular}{@{}ccccc@{}}
\toprule
$p$ & 1 Seg.         & 2 Seg.         & 4 Seg.         & 8 Seg.         \\ \midrule
576 & \cellcolor[HTML]{C0C0C0}34.55 & \cellcolor[HTML]{C0C0C0}34.76 & \cellcolor[HTML]{C0C0C0}21.15 & \cellcolor[HTML]{C0C0C0}34.47 \\
288 & 44.13          & 36.02          & 36.33          & 32.40          \\
144 & 41.66          & 38.16          & 38.06          & 39.93          \\
64  & 37.39          & 39.85          & 36.57          & 37.59          \\
32  & \textbf{45.16} & 41.30          & 43.49          & 38.90          \\
16  & 43.85          & \textbf{44.56} & 41.86          & 39.99          \\
8   & 38.50          & 38.94          & \textbf{44.58} & \textbf{42.38} \\
4   & 43.91          & 42.44          & 41.60          & 41.88          \\ \midrule
$r$ & .39            & .80            & .79            & .86            \\ \bottomrule
\end{tabular}
\end{table}

Table~\ref{tab:spatial-compression} shows the results. We observe a consistent positive correlation between spatial compression rate and downstream accuracy, with the correlation becoming stronger as the number of segments increases, i.e., as segments shorten.

This trend can be understood by examining how spatial compression operates in \espresso{}. Since the spatial compressor is applied after each segment's encoded frames have been temporally pooled, it targets redundancy across spatial locations within these pooled features. Large, relatively static regions, such as sky, court lines, or crowd backgrounds, often persist across the temporally pooled representations of a segment but contribute little to the understanding of events, especially in benchmarks focused on temporal understanding like EgoSchema. Passing these uncompressed features may saturate the LLM's input with low-utility information, making it harder to focus on salient content. In contrast, higher spatial compression encourages \espresso{} to discard such irrelevant regions and retain denser, more informative features. As the number of segments increases, the total token budget grows, and without sufficient spatial compression, much of the additional capacity may be consumed by redundant background content. Hence, spatial compression becomes increasingly important with shorter segments.

These results support the hypothesis that \textbf{increasing the spatial information density of each video token enhances long-form video understanding by discarding redundant spatial details}.




\subsection{Temporal Compression}
\label{subsec:temporal-compression}

We now evaluate the impact of temporal compression on the long-form video understanding capabilities of a VLM. To isolate this effect, we fix the number of sampled frames to 128, remove the spatial compressor from \espresso{}, and vary $t$, the number of temporal query tokens, to control the temporal compression rate. As in the spatial case, we define the temporal compression rate as $-\log(t)$.

\begin{table}
\centering
\caption{EgoSchema accuracy for different temporal compression rates in the default setting. The first row indicates the number of frames per segment, with the total number of sampled frames fixed at 128. $t$ is the number of temporal compression query tokens. Bold entries indicate the highest accuracy in each column. Gray cells indicate no temporal compression. $r$ is the Pearson correlation coefficient between $-\log(t)$ and accuracy.}
\label{tab:temporal-compression}
\begin{tabular}{@{}ccccc@{}}
\toprule
$t$ & 128 frames                    & 64 frames                     & 32 frames                     & 16 frames                     \\ \midrule
128 & \cellcolor[HTML]{C0C0C0}34.25 & N/A                           & N/A                           & N/A                           \\
64  & 39.22                         & \cellcolor[HTML]{C0C0C0}30.59 & N/A                           & N/A                           \\
32  & 35.58                         & 34.17                         & \cellcolor[HTML]{C0C0C0}21.23 & N/A                           \\
16  & 33.39                         & 33.85                         & 21.47                         & \cellcolor[HTML]{C0C0C0}33.83 \\
8   & \textbf{45.28}                & 33.09                         & \textbf{32.10}                & \textbf{43.43}                \\
4   & 37.05                         & \textbf{36.97}                & 24.61                         & 38.38                         \\ \midrule
$r$ & .37                           & .81                           & .53                           & .47                           \\ \bottomrule
\end{tabular}
\end{table}

Table~\ref{tab:temporal-compression} shows a generally positive correlation between temporal compression rate and accuracy, with $r$ values ranging from moderate to strong depending on segment size. However, the correlation here is \textbf{weaker and less consistent than for spatial compression} (Table~\ref{tab:spatial-compression}). Notably, unlike spatial compression, the correlation strength peaks at an intermediate segment length (64 frames) and weakens for both shorter and longer segments.

This trend may be explained by how temporal compression operates in \espresso{}. The temporal compressor is applied after each segment’s encoded frames have already been spatially pooled by a learned spatial pooler, which tends to emphasize high-entropy regions most relevant to video understanding. As a result, the temporally aligned tokens it receives are already information-dense, especially in shorter segments where high-entropy regions are likely to persist across adjacent frames. Moreover, in temporally focused benchmarks like EgoSchema, retaining more temporal detail may further improve performance. Consequently, while temporal compression is still beneficial, \textbf{its gains are modest, reflecting the lower redundancy in temporal features}.

These findings suggest that spatial compression is the primary lever for improving token efficiency, whereas the impact of temporal compression depends more on the segment length and the informativeness of the temporally sampled content.




\subsection{Segments and Spatio-Temporal Compression}
\label{subsec:segments}

\begin{table}
\centering
\caption{Results for the spatial and temporal compression evaluation on EgoSchema in both the default and needle-in-a-haystack (NH) settings. The first number in the Frames column denotes the total number of sampled frames, and the second number refers to the number of frames per segment. The highest EgoSchema accuracy scores are in bold. Pearson correlation coefficients ($r$) between the number of segments and EgoSchema accuracy are also reported.}
\label{tab:spatial-temporal-compression}
\begin{tabular}{@{}lccc@{}}
\toprule
Model                       & Frames  & Default       & NH             \\ \midrule
\multirow{4}{*}{\espressoe} & 128/128 & 40.17         & 32.98          \\
                            & 128/64  & 41.66         & 36.21          \\
                            & 128/32  & 42.18         & 38.02          \\
                            & 128/16  & 50.41         & \textbf{45.08} \\ \midrule
                            & $r$     & 0.97          & 0.99           \\ \midrule
Tarsier (7B)                & 16/16   & 49.9          & 37.49          \\
VideoLLaMA2 (7B)          & 16/16   & \textbf{53.1} & 43.25          \\ \bottomrule
\end{tabular}
\end{table}

Building on insights from Sections~\ref{subsec:spatial-compression} and~\ref{subsec:temporal-compression}, we hypothesize that \espresso{} benefits most from processing a large number of frames divided into fine-grained segments, enabled by aggressive spatial and temporal compression. Intuitively, a longer input video provides more visual context, and segmenting it allows the projector to extract information from localized temporal regions. Compression serves as a selective bottleneck, encouraging \espresso{} to distill salient features from each segment while maintaining a fixed token budget.

To test this hypothesis, we fix the total number of sampled frames at 128 and set both the spatial and temporal query counts ($p$ and $t$) to 4. We then vary the number of segments (i.e., the number of frames per segment) and evaluate the resulting models on both the default and NH-EgoSchema settings. We also compare against two 7B-scale baselines: Tarsier~\citep{wang2024tarsier}, which uses an MLP-based projector, and VideoLLaMA2~\citep{cheng2024videollama}, which uses a pooling-based projector.

As shown in Table~\ref{tab:spatial-temporal-compression}, performance improves consistently as the number of segments increases, with Pearson correlation coefficients of 0.97 and 0.99 on the default and NH variants of EgoSchema, respectively. Notably, when configured with 128 frames and 8 segments, \espresso{} outperforms both Tarsier and VideoLLaMA2 on NH-EgoSchema, and comes close to their performance on the default setting despite using significantly less training data.

These results highlight the benefit of combining aggressive spatial-temporal compression with segmented video processing. \textbf{Segment-wise compression allows \espresso{} to scale to long videos and extract dense, high-quality features, achieving competitive performance with minimal supervision.}

\section{Experiment Compute Resources}
\label{sec:compute-resources}
Each instruction tuning training run was performed on a node with eight 48GB A100 GPUs. The training time depended on the number of sampled frames and segments with the smallest setting of 8 frames and 1 segment taking 3 hours and the largest setting of 128 frames with 8 segments taking 16 hours.

The alignment stage with Panda-70M was performed on a node with eight 48GB H100 GPUs for about a day and a half.

\section{Limitations}
\label{sec:limitations}

While our results validate the viability of fixed-length projectors for long-form and streaming video understanding, several limitations remain.

First, our current training strategy does not yet fully exploit segment-aware pretraining during the initial alignment stage, which may limit the model’s ability to capture long-range temporal dependencies across segments. Incorporating structured objectives or hierarchical pretraining may further enhance the temporal modeling capabilities of \espresso{}.

Second, fixed-length compression introduces an inherent information bottleneck. Although this design improves scalability and efficiency, it can underperform in tasks that require fine-grained temporal reasoning or high-fidelity spatial details, particularly when subtle visual cues must be preserved across time.

Third, fair comparison with prior work remains a challenge due to the heterogeneity of training data across models. Many existing VLMs are trained on vast, privately curated video-text corpora, often unavailable to the broader research community. This discrepancy in data volume, diversity, and accessibility makes it difficult to isolate architectural improvements from advantages conferred by scale or proprietary data.

Addressing these limitations may require both architectural advances and broader community efforts toward standardized benchmarks, open datasets, and transparent reporting of training regimes.

\section{Broader Impacts}
\label{sec:broader-impacts}

This work introduces a fixed-length spatiotemporal compression approach for vision-language models (VLMs), aiming to improve scalability and long-form video understanding. While our method is not directly evaluated in streaming settings, its efficient representation design may support future applications in real-time and resource-constrained environments. Potential positive impacts include enabling more capable assistive agents and expanding access to video understanding in low-bandwidth or edge settings. However, as with other general-purpose VLMs, risks include misuse for surveillance or misaligned automated decisions. We encourage future work to consider appropriate safeguards and deployment oversight.

\end{document}